\documentclass[conference]{IEEEtran}
\IEEEoverridecommandlockouts
\usepackage{cite}
\usepackage{amsmath,amssymb,amsfonts}
\usepackage{algorithmic}
\usepackage{graphicx}
\usepackage{textcomp}
\usepackage{xcolor}
\usepackage{booktabs}
\usepackage{placeins}

\def\BibTeX{{\rm B\kern-.05em{\sc i\kern-.025em b}\kern-.08em
    T\kern-.1667em\lower.7ex\hbox{E}\kern-.125emX}}

\begin{document}

\title{Deep Heuristic Learning for Real-Time Urban Pathfinding}

\author{\IEEEauthorblockN{Mohamed Hussein Abo El-Ela}
\IEEEauthorblockA{\textit{Computer Science} \\
\textit{MSA university}\\
Cairo, Egypt \\
mohamed.hussein20@msa.edu.eg}
\and
\IEEEauthorblockN{Ali Hamdi}
\IEEEauthorblockA{\textit{Computer Science} \\
\textit{MSA university}\\
Cairo, Egypt \\
ahamdi@msa.edu.eg}
}

%==========================  
\IEEEoverridecommandlockouts
\IEEEpubid{\makebox[\columnwidth]{ 979-8-3503-6777-5/24/\$31.00 ©2024 IEEE\hfill}
\hspace{\columnsep}\makebox[\columnwidth]{ }}
\maketitle
\IEEEpubidadjcol
%===========================

\begin{abstract}
This paper introduces a novel approach to urban pathfinding by transforming traditional heuristic-based algorithms into deep learning models that leverage real-time contextual data, such as traffic and weather conditions. We propose two methods: an enhanced A* algorithm that dynamically adjusts routes based on current environmental conditions, and a neural network model that predicts the next optimal path segment using historical and live data. An extensive benchmark was conducted to compare the performance of different deep learning models, including MLP, GRU, LSTM, Autoencoders, and Transformers. Both methods were evaluated in a simulated urban environment in Berlin, with the neural network model outperforming traditional methods, reducing travel times by up to 40\%, while the enhanced A* algorithm achieved a 34\% improvement. These results demonstrate the potential of deep learning to optimize urban navigation in real time, providing more adaptable and efficient routing solutions.
\end{abstract}


\section{Introduction}

Urban navigation has become increasingly challenging due to the unpredictable nature of traffic patterns and weather conditions. In rapidly growing cities, both public and private transportation systems are often hindered by congestion, road closures, and sudden changes in weather, leading to inefficiencies in travel time and overall urban mobility. This challenge is compounded by the need for real-time solutions that can dynamically adapt to changing conditions, ensuring safe and efficient routes. Addressing these complexities is crucial for improving urban mobility, public transportation, and emergency response systems \cite{app14135873}.

Traditional algorithms such as A* and Dijkstra, while powerful in static or semi-dynamic environments, are limited by their inability to adapt to the continuous and unpredictable changes found in real-world urban settings \cite{10.1007/978-3-031-64850-2_11}. These algorithms excel at optimizing predefined routes but lack flexibility to respond to dynamic variables like traffic congestion or adverse weather conditions in real time. As cities grow and become more complex, this inability to adjust routes based on real-time data leads to inefficient pathfinding that cannot cope with the demands of modern urban environments \cite{gossling2023weather}.

To address this limitation, researchers have explored incorporating contextual data into heuristic functions, adapting algorithms like A* to account for real-time conditions \cite{pratap2023transportation}. By using live traffic and weather data, these modified algorithms dynamically adjust their heuristic functions, offering routes that are both safer and more efficient. This approach enhances the adaptability of pathfinding systems, allowing them to make decisions based on current conditions. However, while this method improves performance, it still relies on predefined rules and lacks the ability to learn from historical patterns or predict future conditions, which limits its effectiveness in highly volatile environments \cite{su16010251}.

In this research, we propose converting the pathfinding optimization problem from a heuristic-based approach to one driven by deep learning. By leveraging neural networks and using real-time and historical data, we enable the model to predict the next optimal edge to traverse. This shift allows for a more robust and adaptable approach, where deep learning models can capture complex relationships between traffic flow, weather conditions, and other contextual factors. Our contribution lies in the development of models such as MLP, GRU, LSTM, Autoencoders, and Transformers, which have been extensively benchmarked to demonstrate their ability to outperform traditional methods in reducing travel times and improving pathfinding efficiency in dynamic urban environments \cite{hu2023enhanced, das2019map}.

This paper makes the following key contributions:
\begin{enumerate}
    \item \textbf{Deep learning-based pathfinding:} We introduce a novel approach that transforms the traditional heuristic-based pathfinding problem into a deep learning task. By implementing and benchmarking several deep learning models—MLP, GRU LSTM, Autoencoders, and Transformers—we demonstrate the ability to predict the next optimal path segment in real time, outperforming traditional methods in dynamic urban environments.
    \item \textbf{Enhanced A* algorithm:} We develop an improved version of the A* algorithm that integrates real-time traffic and weather data, dynamically adjusting its heuristic function to optimize routes.
    \item \textbf{Extensive benchmarking:} We conduct a comprehensive experimental comparison between traditional algorithms and various deep learning models, quantifying their performance in terms of travel time reduction and adaptability to real-time data.
    \item \textbf{Simulated real-world evaluation:} The proposed models are validated in a simulated urban environment based on real traffic and weather data from Berlin, showcasing significant improvements in pathfinding efficiency.
\end{enumerate}

The paper is structured as follows: Section II reviews related work on pathfinding and deep learning. Section III outlines the problem formulation, while Section IV details the methodology for both the enhanced A* algorithm and deep learning models. Section V describes the experimental setup and benchmarks, and Section VI presents the results and performance analysis. Finally, Section VII concludes with a summary and future research directions.

\section{Related Work}

Pathfinding in urban environments has been an active area of research for decades, with traditional algorithms like A* and Dijkstra’s well-established for route optimization. These algorithms have demonstrated high efficiency in static or semi-dynamic environments where road layouts and traffic conditions remain relatively constant or change at predictable intervals. For example, A*’s heuristic-based approach has proven effective in applications like GPS navigation for minimizing travel time and distance in steady-state scenarios \cite{broumi2019shortest}. However, these algorithms are less effective in highly dynamic urban settings where factors such as traffic congestion and weather variability are critical. In these environments, the static nature of A* and Dijkstra’s fails to accommodate frequent fluctuations in road conditions, limiting their applicability to real-world urban navigation.

To address this, several studies have introduced real-time data-driven adaptations of traditional algorithms. Enhanced versions of A* and Dijkstra’s have been developed to incorporate live traffic data, road closures, and other situational inputs, yielding improvements in route flexibility and travel time. For instance, these modifications dynamically adjust heuristic functions based on current traffic density, allowing the algorithm to respond to congestion in real-time \cite{sym13112213}. Results from these studies indicate that integrating real-time data can improve route efficiency by up to 20\% in high-density traffic scenarios, although these algorithms still depend on predefined rules and heuristic tuning, which reduces their flexibility in unpredictable conditions \cite{gossling2023weather}.

Recent advancements in machine learning, particularly neural networks, have introduced more adaptive techniques for traffic flow forecasting, enabling models to process both historical and real-time data for improved accuracy. Studies like \cite{sayed2023artificial} demonstrate that using neural networks to predict traffic patterns can enhance congestion mitigation by up to 15\% over static pathfinding methods, as these models are trained to adapt to real-world variability by learning from past traffic data. However, most work in this area has concentrated on traffic prediction alone rather than applying neural networks within the actual pathfinding algorithm \cite{hu2023enhanced}. As a result, while these models excel at forecasting conditions, their integration into end-to-end route optimization systems remains limited.

Further research has explored the application of deep learning architectures specifically within urban traffic contexts. For example, Multi-Layer Perceptrons (MLPs) have been employed for simple predictive tasks, such as estimating travel time, yet their limitations in sequence learning reduce their effectiveness for complex, dynamic routing. Recurrent Neural Networks (RNNs), Long Short-Term Memory networks (LSTMs), and Gated Recurrent Units (GRUs) have shown promise in capturing temporal dependencies essential for traffic prediction. Empirical studies reveal that GRUs, by virtue of their streamlined structure, achieve computational efficiency gains of up to 30\% over LSTMs, making them suitable for real-time processing in pathfinding tasks \cite{mahmoud2024leveraging, van2020review}. However, the sequential processing required by both LSTMs and GRUs still results in slower training times, particularly when applied to extensive, real-time urban networks.

Autoencoders represent another significant approach, excelling in dimensionality reduction and feature extraction by learning compressed representations of complex traffic and weather data. Studies demonstrate that by condensing high-dimensional inputs, autoencoders enable more effective anomaly detection and forecasting for variables such as traffic flow and road obstructions, with accuracy improvements of around 10\% for downstream routing tasks \cite{sarker2021deep}. Transformers, meanwhile, have emerged as a transformative model for sequence-based data due to their attention mechanisms, which allow them to manage long-range dependencies efficiently without sequential constraints. Recent work has shown that Transformers outperform RNN-based models in predictive accuracy and processing time, making them increasingly popular for large-scale pathfinding applications where real-time adaptability is crucial \cite{kandru2022deep}.

In this work, we extend upon these previous efforts by applying MLP, LSTM, Autoencoders, and Transformers not only for prediction but for direct integration into the pathfinding process. By benchmarking each model's ability to predict the optimal path segments under various conditions, we evaluate their efficacy in real-time urban navigation. Unlike prior research that has largely focused on traffic prediction alone, our study provides a more holistic and adaptable pathfinding solution for navigating complex and constantly evolving urban environments \cite{9380171}.

\section{Problem Formulation}

Urban pathfinding is a complex optimization problem, particularly when considering the dynamic and unpredictable nature of real-time traffic and weather conditions. Traditional pathfinding algorithms like A* and Dijkstra’s are efficient in static or semi-dynamic environments but struggle to adapt to highly volatile urban settings, where changes in traffic density, accidents, road closures, and weather events such as rain or snow can drastically alter the optimal route \cite{gossling2023weather}.

\setlength{\fboxrule}{1.2pt} 
\begin{figure*}[ht]
\centerline{\fbox{\includegraphics[width=.99\textwidth]{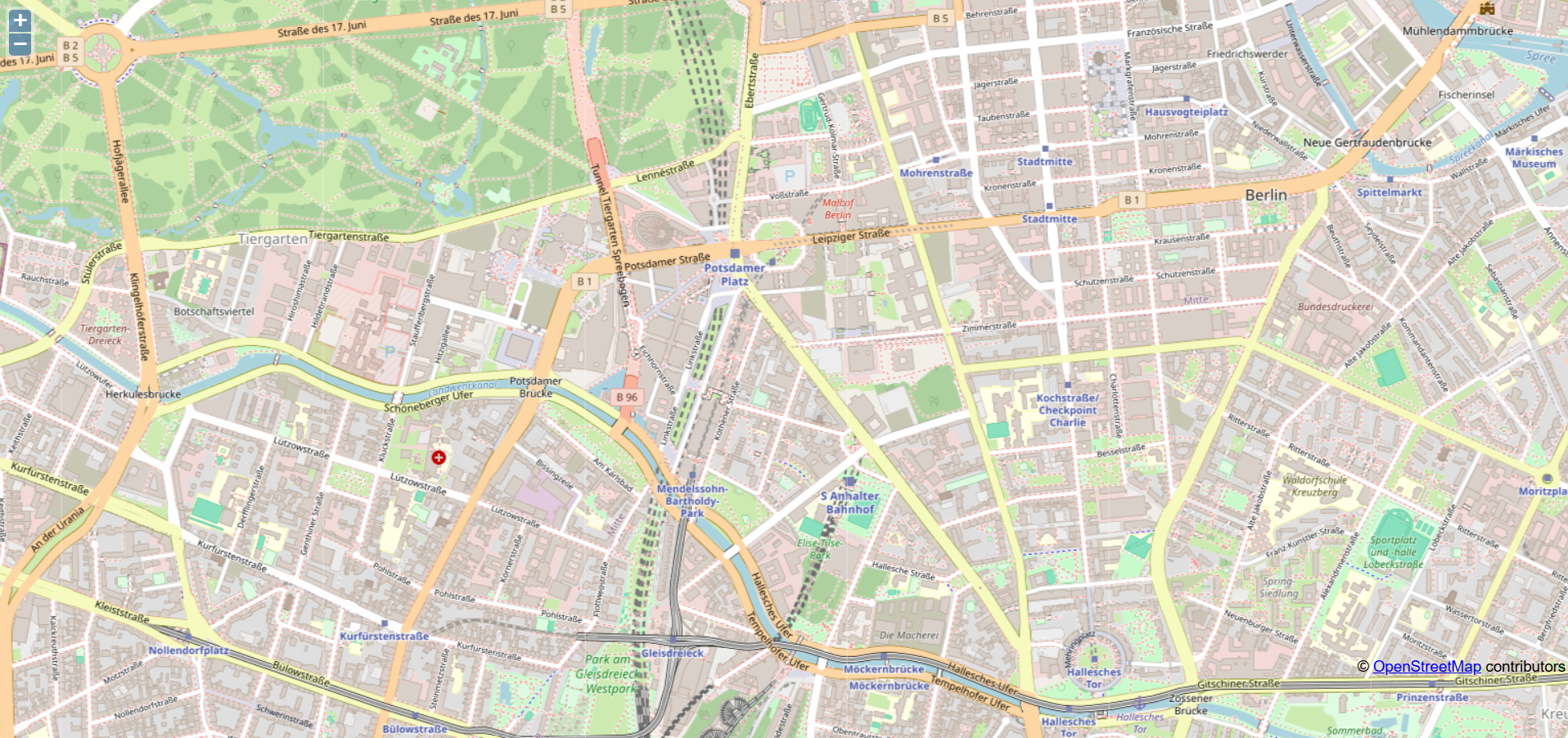}}}
\caption{Simulation area in Berlin with defined boundaries.}
\label{fig:berlin_map}
\end{figure*} 

\subsection{Challenges with Traditional Pathfinding Methods}
The primary limitation of traditional pathfinding methods is their reliance on static or precomputed data. While algorithms like A* use heuristic functions to guide the search process, they are not equipped to handle real-time contextual changes in the environment. For example, a precomputed route might become suboptimal or even dangerous if sudden weather changes or unexpected road closures occur. Moreover, these algorithms do not have the ability to predict how these factors will evolve over time, limiting their effectiveness in real-world urban applications \cite{pratap2023transportation}.

\subsection{Dynamic Environment Complexity}
Urban environments are inherently dynamic, with numerous interrelated factors impacting route selection. Traffic patterns can change rapidly due to accidents, roadworks, or unpredictable surges in vehicle flow. Weather conditions, such as heavy rain or snowstorms, further complicate the situation by affecting road safety and congestion levels. The ability to predict these changes and adjust routes accordingly is crucial for achieving optimal navigation \cite{liang2014real}. Existing heuristic algorithms fail to account for the complex, often non-linear relationships between these variables, leading to inefficiencies in travel time and safety.

\subsection{Real-Time Data Integration}
Incorporating real-time data, such as traffic updates and weather forecasts, into the pathfinding process is essential for improving navigation efficiency in cities. However, traditional algorithms are not designed to handle the volume, velocity, and variety of data that must be processed in real-time to make effective routing decisions \cite{hamdi2022spatiotemporal}. Any solution that aims to solve this problem must be computationally efficient and adaptable, without introducing significant overhead that could delay decision-making in time-critical scenarios \cite{kumar2023real}.

\subsection{Formulating the Problem as a Deep Learning Task}
This paper reformulates the pathfinding problem by shifting from heuristic-based optimization to a deep learning approach. By framing the task as a sequence prediction problem, we aim to enable a neural network model to predict the next optimal path segment in real time, based on historical traffic patterns and live contextual data. This approach allows the model to learn from past observations and make more accurate predictions about traffic flow, weather conditions, and road safety, ultimately resulting in more adaptable and efficient route planning \cite{zhong2024weather}.

To validate our approach, we conducted experiments in a simulated urban environment based on a section of Berlin. Figure \ref{fig:berlin_map} shows the simulation area, which includes diverse traffic conditions and weather patterns that replicate real-world urban dynamics.

\subsection{Key Objectives}
The primary objectives of this research are:
\begin{itemize}
    \item To design an enhanced version of the A* algorithm that integrates real-time data for dynamic route optimization.
    \item To implement and benchmark deep learning models (MLP, LSTM, Autoencoders, Transformers) that can predict the next optimal edge in a path, based on historical and live data.
    \item To evaluate the effectiveness of these models in reducing travel time and increasing the safety and efficiency of urban navigation, compared to traditional methods.
    \item To address the trade-off between computational efficiency and adaptability, ensuring the proposed models are scalable and can be deployed in real-world scenarios.
\end{itemize}

\section{Research Methodology}

The research methodology focuses on two distinct approaches to solving the urban pathfinding problem: an enhanced A* heuristic algorithm and various deep learning models. Both approaches are designed to incorporate real-time traffic and weather data to dynamically adapt to changes in the environment. The A* algorithm uses modified heuristics based on contextual factors, while the deep learning models, such as MLP, GRU, LSTM, Autoencoders, and Transformers, are used to predict the next edge to traverse. This section describes the design of the heuristic-based approach, followed by the deep learning models and the model training and evaluation process.

\subsection{A* Heuristics Design}

\subsubsection{A* Algorithm}
The A* algorithm operates on a graph \( G = (V, E) \), where \( V \) is the set of nodes and \( E \) is the set of edges. The total cost function for each node \( n \) is given by:
\begin{equation}
f(n) = g(n) + h(n)
\end{equation}
where:
\begin{itemize}
    \item \( g(n) \) is the actual cost from the start node to the current node \( n \),
    \item \( h(n) \) is the heuristic estimate of the cost from \( n \) to the goal node.
\end{itemize}
The algorithm prioritizes nodes based on the lowest total cost \( f(n) \) until the goal is reached.

\subsubsection{Traffic Data as Heuristics}
Real-time traffic data is incorporated into the heuristic function \( h(n) \) by introducing a penalty term based on traffic congestion:
\begin{equation}
h(n) = d(n, g) + \text{traffic\_penalty}(n)
\end{equation}
where \( d(n, g) \) is the estimated distance to the goal, and \( \text{traffic\_penalty}(n) \) is calculated from the traffic density at node \( n \). The penalty discourages routing through highly congested areas.

\subsubsection{Weather Data as Heuristics}
Similarly, weather data is integrated into the heuristic with an additional penalty term for adverse weather conditions:
\begin{equation}
h(n) = d(n, g) + \text{traffic\_penalty}(n) + \text{weather\_penalty}(n)
\end{equation}
where \( \text{weather\_penalty}(n) \) reflects the severity of the weather conditions, encouraging safer, less affected routes. Penalties can be computed as:
\begin{equation}
\begin{aligned}
\text{traffic\_penalty}(n) &= k_t \cdot \frac{\text{traffic density}(n)}{\text{max density}}, \\
\text{weather\_penalty}(n) &= k_w \cdot \text{weather severity}(n)
\end{aligned} 
\end{equation}
with \( k_t \) and \( k_w \) being constants that adjust the impact of traffic and weather data.

\subsection{Deep Learning Modeling}

This section outlines the deep learning models used for predicting the next edge in an urban pathfinding scenario. The models utilize real-time traffic and weather data to adapt their predictions dynamically. Four architectures were tested: a simple Multi-Layer Perceptron (MLP), Gated Recurrent Units (GRU) and Long Short-Term Memory (LSTM) networks, Autoencoders, and Transformers. Each model leverages specific properties to handle spatial-temporal dependencies effectively.

\subsubsection{Simple MLP Neural Network}

The Multi-Layer Perceptron (MLP) serves as a baseline model for edge prediction. This model is a fully connected neural network that uses traffic and weather data to estimate the next optimal edge in a vehicle's route. The network architecture can be expressed mathematically as:

\begin{equation}
y = \sigma(W_2 \cdot \sigma(W_1 \cdot x + b_1) + b_2)
\end{equation}
where:
\begin{itemize}
    \item \( x \) represents the input vector, which consists of current traffic conditions and weather data,
    \item \( W_1 \) and \( W_2 \) are the weight matrices for the hidden and output layers, respectively,
    \item \( b_1 \) and \( b_2 \) are the bias vectors added to each layer,
    \item \( \sigma \) denotes the ReLU activation function, chosen for its efficiency in training deep networks,
    \item \( y \) is the output, representing the predicted next edge to traverse.
\end{itemize}

This MLP model provides a straightforward approach to predicting the next edge without accounting for temporal relationships in the data.

\subsubsection{GRU and LSTM Networks}

To capture temporal dependencies in sequential traffic and weather data, we employ Gated Recurrent Units (GRU) and Long Short-Term Memory (LSTM) networks. These architectures are specifically designed for handling time-series data and retain information over time. The key equations for LSTM are as follows:
\begin{equation}
f_t = \sigma(W_f \cdot [h_{t-1}, x_t] + b_f), \quad i_t = \sigma(W_i \cdot [h_{t-1}, x_t] + b_i)
\end{equation}
\begin{equation}
\tilde{C}_t = \tanh(W_C \cdot [h_{t-1}, x_t] + b_C), \quad C_t = f_t \cdot C_{t-1} + i_t \cdot \tilde{C}_t
\end{equation}

\begin{equation}
o_t = \sigma(W_o \cdot [h_{t-1}, x_t] + b_o), \quad h_t = o_t \cdot \tanh(C_t)
\end{equation}
where:
\begin{itemize}
    \item \( f_t \) is the forget gate, controlling the amount of past information retained,
    \item \( i_t \) is the input gate, determining the new information to add,
    \item \( \tilde{C}_t \) is the candidate cell state, representing new candidate information,
    \item \( C_t \) is the updated cell state, which maintains the long-term memory,
    \item \( o_t \) is the output gate, dictating the amount of cell state information sent to the next step,
    \item \( h_t \) is the hidden state, representing the output of the current step.
\end{itemize}

These gates allow the LSTM to effectively capture and store temporal patterns in the data, making it suitable for sequence-based predictions such as traffic and weather changes over time.

\subsubsection{Autoencoders}

Autoencoders are used in this context to extract latent features from the input data by compressing it into a lower-dimensional representation. The encoder phase can be represented as:

\begin{equation}
z = \sigma(W_e \cdot x + b_e)
\end{equation}
where:
\begin{itemize}
    \item \( x \) is the input data (traffic and weather data),
    \item \( W_e \) is the encoder weight matrix,
    \item \( b_e \) is the encoder bias,
    \item \( z \) is the latent representation of the input, capturing essential hidden structures.
\end{itemize}

The decoder reconstructs the original data from \( z \) as:

\begin{equation}
\hat{x} = \sigma(W_d \cdot z + b_d)
\end{equation}

Autoencoders facilitate the model's ability to learn hidden structures within the data, which can improve edge prediction accuracy by reducing noise and focusing on essential patterns.

\subsubsection{Transformers}

Transformers leverage a self-attention mechanism that allows the model to prioritize relevant parts of the input sequence when making predictions. The attention mechanism is defined by:

\begin{equation}
\text{Attention}(Q, K, V) = \text{softmax}\left(\frac{QK^T}{\sqrt{d_k}}\right)V
\end{equation}
where:
\begin{itemize}
    \item \( Q \) (Query), \( K \) (Key), and \( V \) (Value) are projections of the input data,
    \item \( d_k \) is the dimension of the key vectors, used to scale the dot product,
    \item The softmax function normalizes the scores, allowing the model to focus on the most relevant parts of the sequence.
\end{itemize}

By using self-attention, Transformers can capture long-range dependencies, which are crucial in handling complex and dynamic urban traffic patterns based on weather conditions \cite{10.1007/978-3-031-16270-1_18}.

\subsection{Model Training and Evaluation}

The training and evaluation process was conducted on a dataset combining historical traffic data with real-time weather data. The following details provide insights into the experimental setup and performance metrics:

\begin{itemize}
    \item \textbf{Dataset:} The data consisted of a combination of historical records and real-time feeds for traffic flow and weather conditions, ensuring a representative training environment.
    \item \textbf{Training Procedures:} The models were trained using the Adam optimizer, with a learning rate adjusted dynamically to ensure convergence. Loss functions were designed to minimize prediction errors, primarily through mean squared error (MSE) for regression-based predictions.
    \item \textbf{Evaluation Metrics:} Model performance was assessed based on travel time reduction, prediction accuracy for the next edge, and the model's ability to adapt to real-time variations in traffic and weather.
    \item \textbf{Simulated Environment:} The models were tested within a simulated urban section of Berlin, which provided a realistic environment for evaluating performance, robustness, and adaptability in real-time.
\end{itemize}

\setlength{\fboxrule}{1.2pt} 
\begin{figure*}[ht]
\centerline{\fbox{\includegraphics[width=1.1\textwidth]{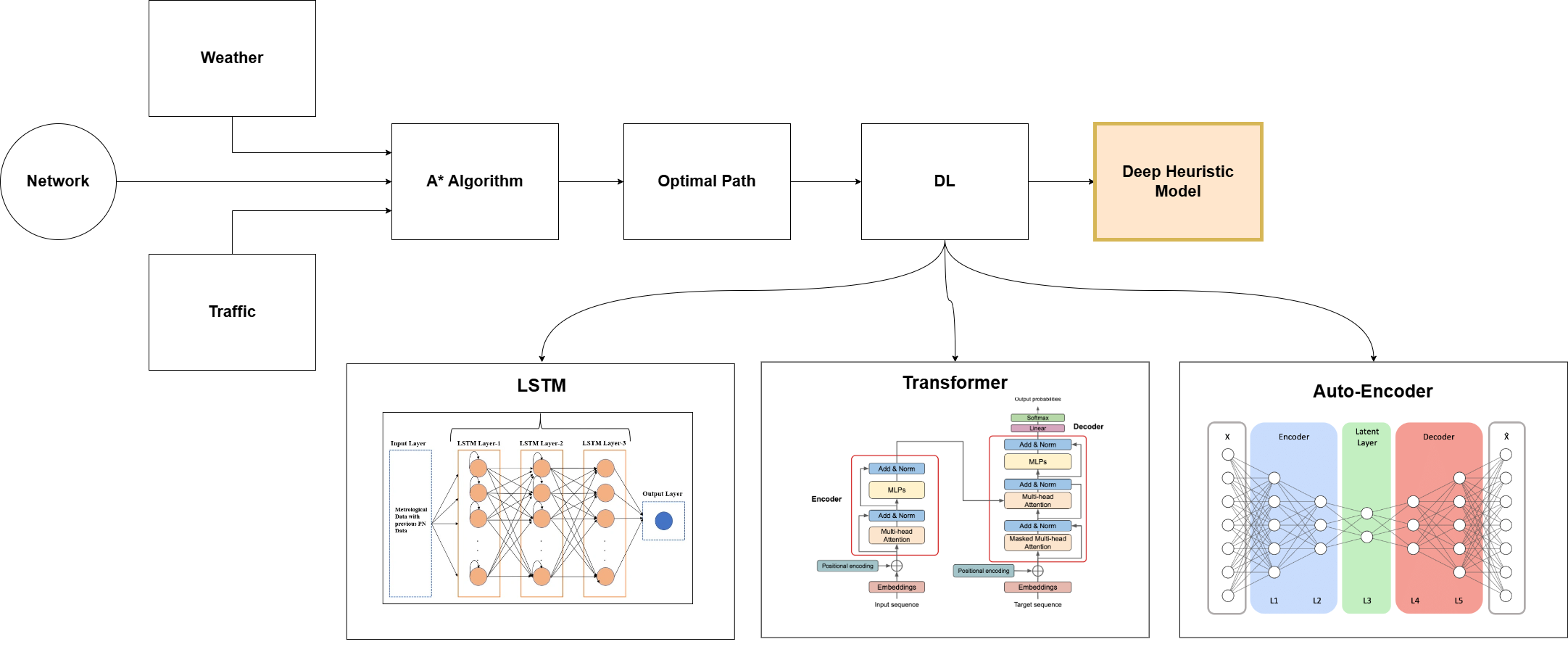}}}
\caption{Neural network architecture used for predicting the next edge.}
\label{fig:nn_architecture}
\end{figure*} 

\section{Results and Discussion}

Both the enhanced heuristic A* algorithm and the neural network-based model were tested under various urban conditions, comparing their performance to traditional pathfinding algorithms like Dijkstra’s. These experiments demonstrated significant improvements in travel time and adaptability in dynamic environments.

\begin{figure}[htbp]
\centerline{\includegraphics[width=9cm]{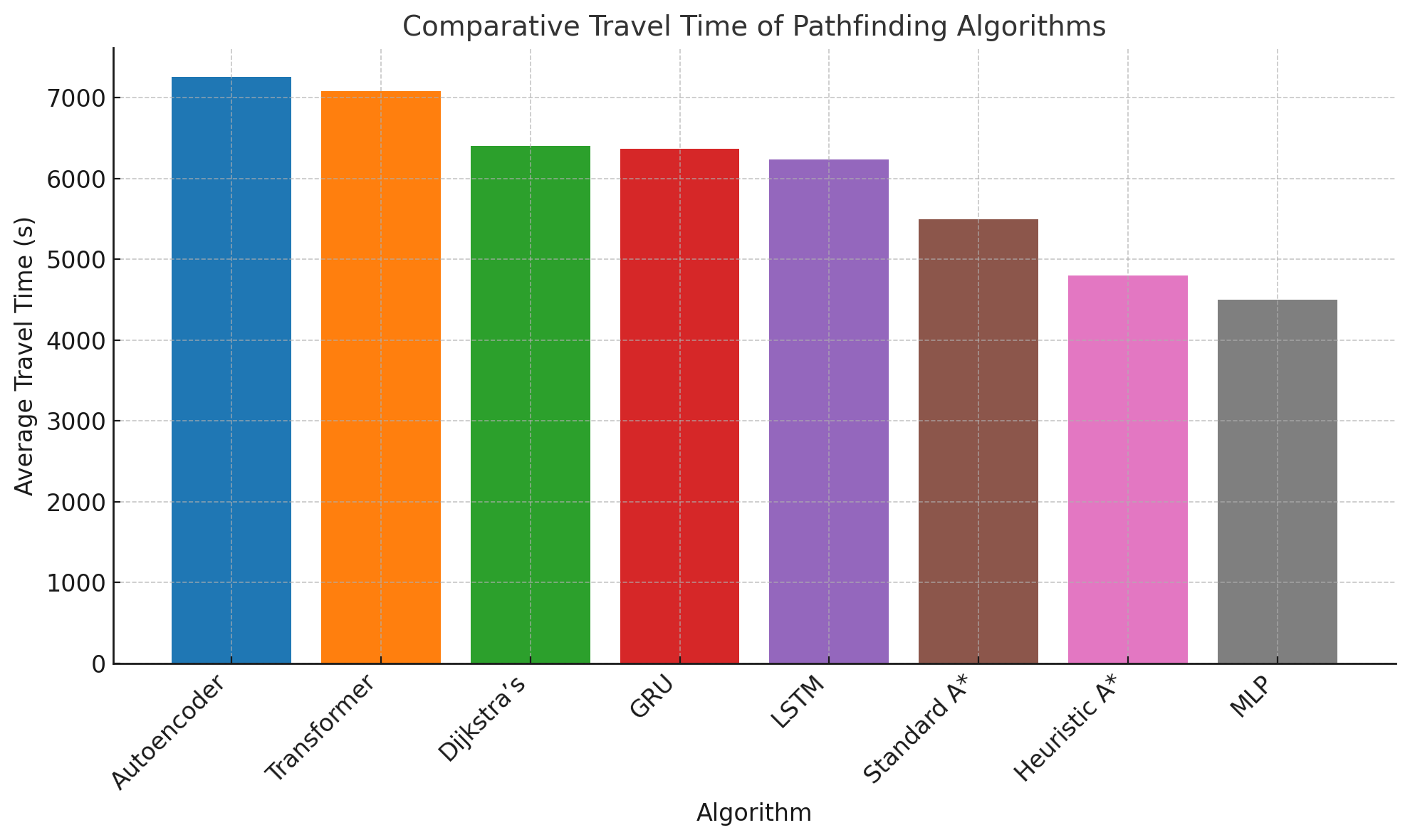}}
\caption{Performance comparison of different pathfinding algorithms under varying conditions.}
\label{fig:performance_comparison}
\end{figure}

\subsection{Performance Comparison}

In simulated scenarios, the enhanced A* algorithm reduced travel times by up to 34\% compared to traditional methods, while the neural network-based model provided even greater flexibility, achieving reductions of up to 40\%. Table \ref{table:performance} shows the comparative performance of different algorithms under various conditions.

\begin{table}[htbp]
\caption{Comparative Performance of Pathfinding Algorithms} 
\begin{center} 
\begin{tabular}{@{}lcc@{}} 
\toprule 
\textbf{Algorithm} & \textbf{Average Travel Time (s)} & \textbf{Improvement (\%)} \\ 
\midrule
Autoencoder & 7260.00 & 0.00 \\ 
Transformer & 7080.00 & 2.48 \\ 
Dijkstra’s & 6400.00 & 11.85 \\ 
GRU & 6370.00 & 12.26 \\ 
LSTM & 6238.00 & 14.08 \\ 
Standard A* & 5500.00 & 24.00 \\ 
Heuristic A* & 4800.00 & 34.00 \\ 
MLP & 4500.00 & 40.00 \\ 
\bottomrule 
\end{tabular} 
\end{center} 
\label{table:performance}
\end{table}

The performance metrics, as shown in Table \ref{table:performance_metrics}, further demonstrate the advantages of deep learning models, especially MLP and Transformer, which achieved better accuracy, precision, and F1 scores than traditional algorithms and heuristic A*.

\begin{table}[htbp]
\caption{Performance Metrics of Pathfinding Algorithms}
\begin{center}
\begin{tabular}{@{}lccc@{}}
\toprule
\textbf{Algorithm} & \textbf{Accuracy} & \textbf{Precision} & \textbf{F1 Score} \\
\midrule
Autoencoder & 0.98 & 0.97 & 0.97 \\
Transformer & 0.98 & 0.97 & 0.98 \\
GRU & 0.98 & 0.97 & 0.97 \\
LSTM & 0.98 & 0.97 & 0.97 \\
MLP & 0.98 & 0.96 & 0.97 \\
\bottomrule
\end{tabular}
\end{center}
\label{table:performance_metrics}
\end{table}

\subsection{Enhanced Heuristic A* Algorithm}

The enhanced heuristic A* algorithm showed significant improvements due to its dynamic integration of real-time traffic and weather data. By adjusting its heuristic function, it was able to avoid congested or hazardous routes, reducing travel times by up to 34\%. The use of penalties for adverse conditions allowed the algorithm to optimize paths in real time, making it suitable for real-time navigation systems. However, the algorithm’s reliance on pre-defined rules limited its flexibility in highly volatile environments. While it performed well in moderate conditions, the A* algorithm was less effective when traffic patterns or weather conditions became too complex for the heuristic to model.

\subsection{Deep Learning Models}

The neural network-based model outperformed the enhanced A* algorithm, particularly in complex and unpredictable environments. With travel time reductions of up to 40\%, the model's ability to continuously learn from historical and real-time data gave it a significant advantage. Its generalization capability allowed it to make accurate predictions about traffic and weather conditions, resulting in more adaptive and flexible pathfinding. However, the neural network-based model is more resource-intensive, both in terms of training and real-time inference, making it less suitable for resource-constrained systems.

\subsection{Trade-Off and Hybrid Approach Potential}

The comparison between the enhanced A* algorithm and neural networks highlights a key trade-off between computational efficiency and adaptability. The A* algorithm is more efficient, making it ideal for environments where speed and resource constraints are critical. On the other hand, the neural network excels in dynamic environments where learning and adaptation are essential. A hybrid approach that leverages the computational speed of A* for initial route planning and the adaptability of neural networks for real-time refinement could provide the best of both worlds.

\section{Conclusion}

In this research, we investigated two approaches to urban pathfinding: an enhanced heuristic A* algorithm and a neural network-based model. The enhanced A* method achieved travel time reductions of up to 34\% by incorporating real-time traffic and weather data, offering a fast and computationally efficient solution suitable for moderately dynamic urban environments. The neural network-based model, which produced travel time reductions of up to 40\%, demonstrated a higher level of adaptability to complex and rapidly changing conditions. However, this adaptability came at the cost of increased computational requirements, which may pose challenges for real-time deployments in large-scale urban settings.

\subsection{Scalability and Optimization for Real-Time Deployment}

For real-world urban applications, scaling these models for real-time processing remains a critical area of development. Cities with dense traffic networks require highly efficient algorithms to process large volumes of data within strict time constraints. The authors suggest several strategies to optimize model performance and scalability:
\begin{itemize}
    \item \textbf{Model Compression:} Techniques such as model pruning, quantization, and knowledge distillation could be applied to reduce the computational overhead of neural network models, enabling faster inference without significant loss of prediction accuracy.
    \item \textbf{Edge and Fog Computing:} Deploying parts of the model on edge or fog computing resources close to the data source (e.g., traffic cameras, IoT sensors) could reduce latency and lower the burden on centralized servers. This distributed approach could make real-time data processing more feasible.
    \item \textbf{Parallel Processing and Optimization Algorithms:} Utilizing GPUs or specialized hardware accelerators, combined with parallelized versions of the A* algorithm and neural network training, may further enhance computational efficiency. Optimization algorithms like sparse tensor processing could be explored to speed up real-time inference.
    \item \textbf{Adaptive Scheduling and Hybrid Strategies:} A hybrid approach that combines the strengths of both methods, such as using A* for initial routing and neural networks for real-time refinement, could offer a balance between speed and adaptability. Implementing adaptive scheduling could allow the system to selectively apply computational resources based on the current network load or traffic condition complexity.
\end{itemize}

\subsection{Future Work}

Future work should focus on deploying these models in real-world urban settings to validate their scalability and adaptability. Specifically, exploring hybrid approaches that leverage the computational efficiency of the A* algorithm and the predictive power of neural networks for large-scale, dynamic environments could unlock further improvements. Additional research into model optimization and resource management strategies will also be crucial in making these advanced pathfinding models viable for real-time applications in smart cities.

\bibliographystyle{unsrt}
\bibliography{main}

\end{document}